\documentclass{article}
\usepackage{spconf,amsmath,graphicx}
\usepackage[T1]{fontenc}
\usepackage{newtxtext,newtxmath}
\usepackage{booktabs}
\usepackage{multirow}
\usepackage{adjustbox}
\usepackage{makecell}
\usepackage[pagebackref,breaklinks,hidelinks]{hyperref}

\title{LiPS: Lightweight Panoptic Segmentation \\ for Resource-Constrained Robotics}

\name{
Calvin Galagain$^{1,2}$,
Martyna Poreba$^{1}$,
François Goulette$^{2}$,
Cyrill Stachniss$^{3}$
}

 \address{
 $^{1}$ Université Paris-Saclay, CEA LIST, F-91120 Palaiseau, France \\
 $^{2}$ U2IS, ENSTA, Institut Polytechnique de Paris, 91120 Palaiseau, France \\
 $^{3}$ University of Bonn, Center for Robotics, Bonn, Germany
 }

\begin{document}
\maketitle

\begin{abstract}

Panoptic segmentation is a key enabler for robotic perception, as it unifies semantic understanding with object{-}level reasoning. However, modern query{-}based models are often computationally and memory intensive, which can make deployment difficult on power{-}constrained robotic platforms.
We propose LiPS (Lightweight Panoptic Segmentation), an efficient query{-}based panoptic segmentation framework that preserves masked transformer decoding while reducing the computational cost of upstream processing. LiPS combines selective feature routing, spatial compression before multi{-}scale fusion, and a shallow reduced{-}width pixel decoder.
Experiments on ADE20K and Cityscapes show that LiPS preserves most panoptic and semantic performance while reducing computation by up to $6.8\times$ and improving embedded throughput by up to $4.5\times$. Additional 15\,W measurements on Jetson AGX Orin confirm lower latency and GPU memory usage, making LiPS suitable for robotic perception pipelines where throughput and memory footprint are prioritized over fine instance boundaries.
\end{abstract}


\section{Introduction}
\label{sec:intro}

Panoptic segmentation brings together semantic and instance segmentation into a single pixel{-}wise representation, distinguishing between \textit{things} (countable objects) and \textit{stuff} (amorphous regions such as road or sky)~\cite{Kirillov2018PanopticS}.
This unified view is particularly valuable in robotics, where perception systems must jointly support global scene understanding and object{-}level reasoning for tasks such as semantic mapping, object{-}aware planning, and instance{-}level data association.

Recent progress has been driven by query{-}based transformers, which cast segmentation as a set prediction problem where learned queries interact with image features through attention to predict masks and categories. Mask2Former~\cite{Cheng2021MaskedattentionMT} has become a widely adopted representative of this paradigm. However, query{-}based segmentation models remain computationally and memory intensive due to high{-}capacity backbones, dense multi{-}scale features, and costly pixel{-}level fusion, causing significant latency and energy consumption~\cite{XIANG2023373}.These demands can be difficult to reconcile with embedded robotic deployment, where perception must often run under limited power, memory, and latency budgets on Jetson{-}class devices or similar accelerators.

To tackle these limitations, we propose LiPS (Lightweight Panoptic Segmentation) as an efficient query{-}based panoptic segmentation method designed for resource{-}constrained deployment. LiPS preserves the masked transformer decoder of Mask2Former while streamlining the upstream feature pathway through: (i) a compact hierarchical encoder, (ii) selective feature routing, and (iii) a shallow reduced{-}width deformable{-}attention pixel decoder. This design targets the dominant computational stages without redesigning the query decoder.
Unlike pruning, quantization, or distillation methods that compress model parameters or token representations, or lightweight backbone replacements that mainly reduce encoder cost,  LiPS structurally reduces the dense multi{-}scale feature pathway while keeping the reasoning module
intact. Experiments on ADE20K~\cite{zhou2017scene} and Cityscapes~\cite{Cordts2016Cityscapes}, conducted on an NVIDIA Jetson AGX Orin with TensorRT FP16 optimization, show that LiPS reduces computation by up to $6.8\times$ and improves throughput by up to $4.5\times$ compared to Mask2Former, while preserving most of the panoptic quality.

\section{Related Work}
\label{sec:related}

\begin{figure*}[t]
    \centering
    \includegraphics[width=\textwidth]{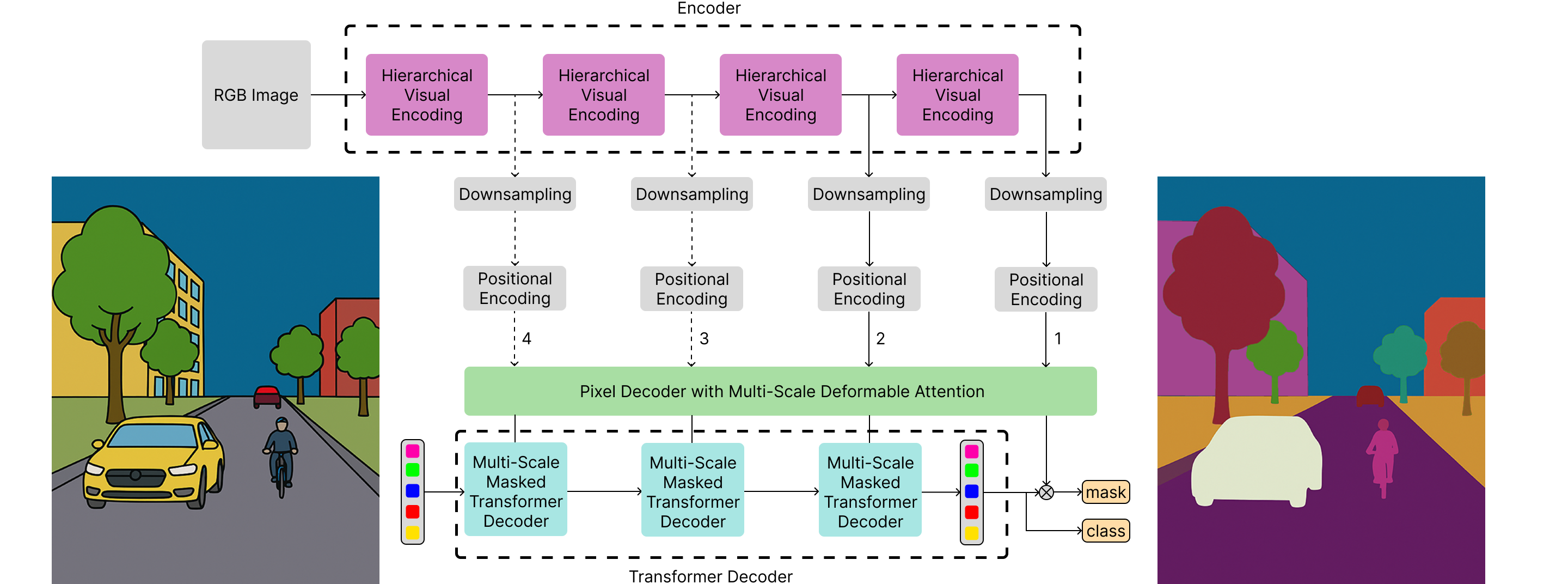}
    \caption{LiPS architecture. Encoder produces a four{-}level feature hierarchy. A routing step selects a subset of levels (1{-}4), which are downsampled with strided convolutions, enriched with sine positional encodings, and fused through a shallow deformable{-}attention pixel decoder. A lightweight top{-}down FPN exposes three mask{-}feature scales for the masked transformer decoder, which predicts class labels and panoptic masks from a fixed set of queries. By default, two routed levels (1 and 2) are used for embedded deployment. Dashed arrows denote skipped levels; solid arrows indicate active information flow.}
    \label{fig:architecture}
\end{figure*}

Query{-}based transformers have profoundly influenced panoptic segmentation by framing it as a set prediction problem driven by learned queries. Following DETR{-}style set prediction and Deformable DETR~\cite{Zhu2020DeformableDD}, MaskFormer~\cite{Cheng2021PerPixelCI} and, most notably, Mask2Former~\cite{Cheng2021MaskedattentionMT} established masked transformer decoding over multi{-}scale feature representations as a prevailing design choice. Building on this paradigm, later works such as kMaX-DeepLab~\cite{Yu2022kMaXDeepLabKM} and OneFormer~\cite{Jain2022OneFormerOT}, and open-vocabulary panoptic segmentation \cite{chen2024open}  introduced improvements in query formulation, decoder design, training strategies, or task unification across segmentation settings.

Across these approaches, query{-}based decoding assigns masks and semantic labels from backbone features.

Lightweight Vision Transformers such as PVT~\cite{Wang2021PyramidVT}, SegFormer~\cite{xie2021segformersimpleefficientdesign}, or AFFormer~\cite{Dong2023HeadFreeLS} demonstrated strong accuracy{-}efficiency trade{-}offs in semantic segmentation. In contrast, efficiency in panoptic segmentation received comparatively less attention, with only a few works adapting lightweight backbones to query{-}based frameworks, e.g., Panoptic SegFormer~\cite{li2021panoptic} or simplifying query{-}based designs like EoMT~\cite{kerssies2025eomt}.

\section{Our Approach}
\label{sec:method}

\subsection{Baseline Analysis}

We adopt Mask2Former~\cite{Cheng2021MaskedattentionMT} as our point of departure due to its strong generalization across semantic, instance, and panoptic segmentation. Its architecture can be decomposed into three main components: a hierarchical encoder that extracts multi{-}scale representations, a pixel decoder that fuses these features, and a masked transformer decoder that performs query{-}based reasoning. Despite its conceptual elegance, Mask2Former remains computationally demanding even with lightweight backbones such as ResNet{-}50. Stage{-}wise profiling of the model reveals a strong imbalance in computational cost, with the encoder accounting for 31.6\% of the total GFLOPs, the pixel decoder dominating at 53.7\%, and the masked transformer decoder contributing only 5.5\%, while the remaining computation is attributed to the final prediction head. This analysis underscores a structural inefficiency where most of the computational budget is allocated to dense multi{-}scale feature extraction and fusion rather than to query{-}based reasoning.

\subsection{Methodology}

Based on this observation we propose LiPS, a lightweight redesign of Mask2Former that preserves its core architectural principles while directly addressing its stage{-}wise computational imbalance.
Our method relies on three complementary ideas: (i) selecting only a subset of encoder stages, (ii) compressing spatial representations before multi{-}scale fusion, and (iii) simplifying the pixel decoder in depth and width. The general architecture of LiPS is illustrated in Figure ~\ref{fig:architecture}.

\textbf{Encoder and Feature Routing.} LiPS employs AFFormer~\cite{Dong2023HeadFreeLS} as a compact hierarchical encoder producing four feature levels with progressively reduced spatial resolution. In contrast to conventional architectures that propagate the full feature pyramid, LiPS employs a routing mechanism to forward only a subset of encoder levels to the pixel decoder. By default, the two highest{-}resolution stages are selected, as they retain most of the spatial cues required for small{-}object delineation while remaining computationally inexpensive. To further reduce the token budget, the routed feature maps are spatially compressed prior to fusion using strided $3\times3$ convolutions. Each selected encoder level is uniformly downsampled with progressively smaller stride factors from coarse to fine resolutions ($\times3$, $\times2$, $\times2$, $\times2$), reducing both spatial resolution and channel dimensionality.
This operation lowers both spatial resolution and channel dimensionality while preserving the hierarchical structure of the features.

\textbf{Lightweight Pixel Decoder.} The routed features are fused through a top{-}down deformable{-}attention pixel decoder~\cite{Zhu2020DeformableDD}. LiPS introduces two efficiency{-}oriented design changes. First, the depth of the deformable transformer encoder is reduced from 6 to 3 layers. Second, the channel width of the convolutional adapters is reduced from 256 to 128. Since convolutional cost scales quadratically with channel dimension, this change decreases both parameters and computational complexity by a factor of four. Once fused, the features are passed through a lightweight FPN{-}style pathway with bilinear upsampling and $1{\times}1$ lateral connections, producing compact mask features at $1/8$, $1/16$, and $1/32$ input resolutions.

\textbf{Transformer Decoder.} The masked transformer decoder of Mask2Former is kept unchanged. As shown in the computational analysis, this component is not a bottleneck and has limited impact on latency in embedded settings. Preserving it allows LiPS to maintain the original interaction between queries and masks while isolating the effect of upstream restructuring. Queries interact through self{-}attention and masked cross{-}attention over the multi{-}scale mask features to predict class labels and masks, from which final masks are produced via projection and sigmoid activation.

\section{Experiments}
\label{sec:experiments}

\subsection{Experimental Setup}

We evaluate LiPS on ADE20k~\cite{zhou2017scene} and Cityscapes~\cite{Cordts2016Cityscapes} to quantify the accuracy{-}efficiency trade{-}offs attained by modifying the upstream computational structure while keeping the query{-}based decoder unchanged. Input image size is fixed to $640{\times}640$ for ADE20K and $512{\times}1024$ for Cityscapes.

All experiments are conducted on an NVIDIA Jetson AGX Orin platform equipped with 64,GB LPDDR5 memory and an Ampere GPU featuring 2048 CUDA cores and 64 Tensor Cores. Unless otherwise stated, experiments are conducted in MaxN mode (60,W), with additional results reported in 15,W mode for power-constrained settings.

Panoptic quality is evaluated using Panoptic Quality (PQ), semantic segmentation using mean Intersection-over-Union (mIoU), and instance segmentation using Average Precision (AP). All metrics are reported in percentage (\%).

Efficiency is reported in GFLOPs per forward pass (computed with fvcore's \texttt{FlopCountAnalysis})~\footnote{https://github.com/facebookresearch/fvcore}, number of parameters, throughput in frames per second (FPS), latency, and peak memory usage with batch size 1. All evaluations use single{-}scale inference without test{-}time augmentation.

To isolate the effect of the restructuring, the masked transformer decoder configuration is kept identical across all variants. Models are implemented in PyTorch, trained following the standard Mask2Former schedule, exported to ONNX, and optimized using TensorRT in FP16 to leverage the Orin's Tensor Cores.

\subsection{Results and Efficiency}

\subsubsection{Main performance}

Tables~\ref{tab:ade20k_ablation_full} and~\ref{tab:cityscapes_ablation} compare LiPS with Mask2Former{-}R50. On ADE20K, LiPS (full) reaches 36.4$\%$\,PQ and 45.3$\%$\,mIoU at 36.8\,GFLOPs, compared to 39.6$\%$\,PQ and 46.0$\%$\,mIoU at 147.2\,GFLOPs for the baseline.

This corresponds to a $4.0\times$ compute reduction with only a 3.2 PQ drop and a 0.7 mIoU drop. LiPS (full) also reduces the model size from 44.0M to 19.8M parameters while increasing embedded throughput from 7.1 to 10.7 FPS.

On Cityscapes, the same trend is observed at higher input resolution. LiPS (full) obtains 55.3\%\,PQ at 121.5\,GFLOPs, versus 62.1\%\,PQ at 527.4\,GFLOPs for Mask2Former{-}R50, corresponding to more than a $4.3\times$ reduction in computation. In addition, LiPS (full) reduces the parameter count from 44.0M to 19.7M and improves throughput from 2.4 to 4.5 FPS.

\begin{table}[h]
\centering
\caption{LiPS performance on ADE20K across different routed encoder levels. }
\resizebox{0.5\textwidth}{!}{
\begin{tabular}{lcccccc}
\toprule
\textbf{Model Variant} & \textbf{PQ(\%)} & \textbf{mIoU(\%)} & \textbf{AP(\%)} & \textbf{GFLOPs} & \textbf{Params (M)} & \textbf{FPS} \\
\midrule
Mask2Former-R50 & 39.6 & 46.0 & 26.5 & 147.2 & 44.0 & 7.1 \\
LiPS (1 level)     & 34.8 & 43.9 & 19.8 & 24.6  & 20.1 & 18.3 \\
LiPS (2 levels)    & 35.8 & 44.0 & 21.4 & 26.4  & 20.0 & 17.5 \\
LiPS (3 levels)    & 36.1 & 44.4 & 22.0 & 28.4  & 19.9 & 15.7 \\
LiPS (full) & 36.4 & 45.3 & 22.1 & 36.8 & 19.8 & 10.7 \\
\bottomrule
\end{tabular}
}
\label{tab:ade20k_ablation_full}
\end{table}

\begin{table}[h]
\centering
\caption{LiPS performance on Cityscapes across different routed encoder levels.}
\resizebox{0.5\textwidth}{!}{
\begin{tabular}{lcccccc}
\toprule
\textbf{Model Variant} & \textbf{PQ(\%)} & \textbf{mIoU(\%)} & \textbf{AP(\%)} & \textbf{GFLOPs} & \textbf{Params (M)} & \textbf{FPS} \\
\midrule
Mask2Former-R50 & 62.1 & 77.5 & 37.3 & 527.4 & 44.0 & 2.4 \\
LiPS (1 level)     & 54.1 & 72.7 & 24.9 & 77.4  &  19.9 & 10.8 \\
LiPS (2 levels)    & 54.5 & 73.3 & 25.9 & 84.2  & 19.9 & 10.7 \\
LiPS (3 levels)    & 55.3 & 73.0 & 25.6 & 91.7  & 19.8 & 8.5 \\
LiPS (full) & 55.3 & 71.6 & 26.9 & 121.5 & 19.7 & 4.5 \\
\bottomrule
\end{tabular}
}
\label{tab:cityscapes_ablation}
\end{table}

\subsubsection{Effect of Encoder Routing}

We analyze the accuracy-efficiency trade-off of LiPS through LiPS (1 level), LiPS (2 levels), and LiPS (3 levels), where deeper routing progressively increases computational cost and representational capacity.

The one{-}level model gives the largest speedup and lowest compute, but also the strongest AP drop, indicating that aggressive removal of scales mainly harms instance delineation. Moving from one to two routed levels partially recovers this loss with limited overhead. On ADE20K, PQ increases from 34.8\% to 35.8\% while GFLOPs rise from 24.6 to 26.4. On Cityscapes, PQ improves from 54.1\% to 54.5\% with GFLOPs increasing from 77.4 to 84.2.
Adding a third or fourth routed level yields diminishing returns relative to the extra feature traffic. For instance, on ADE20K the full model improves over LiPS (2 levels) by only 0.6 points PQ, while increasing compute from 26.4 to 36.8\,GFLOPs and reducing FPS from 17.5 to 10.7. This supports the use of two routed levels as the default embedded configuration. Across both datasets, AP is the most sensitive metric to routing depth, whereas PQ and mIoU remain comparatively stable, confirming that the main cost of compression is localized around small and thin instances rather than global semantic consistency.

\subsubsection{Scaling with Image Resolution}

LiPS scales more favorably than Mask2Former as image resolution increases because it suppresses dense upstream multi{-}scale feature growth before fusion. The compute reduction of LiPS (full) over Mask2Former increases from $2.2{\times}$ at $256^2$ to $3.8{\times}$ at $640^2$ and $4.5{\times}$ at $2048^2$. The two{-}level configuration is even lighter, requiring only 11.6\,GFLOPs at $512^2$ and 15.7\,GFLOPs at $640^2$, compared to 52.2 and 79.7\,GFLOPs for Mask2Former, respectively. This trend is important for robotics, where camera streams are often processed at high resolution and where the cost of dense feature pyramids grows quickly with the number of pixels. Overall, the advantage of LiPS becomes stronger as the input resolution increases.

\subsubsection{Ablation studies}

Before integrating the simplified pixel decoder into LiPS, we validate its design changes directly on Mask2Former{-}R50 to isolate each modification under a fixed backbone and decoder. Table~\ref{tab:enc_layers_ablation} shows that reducing the number of deformable{-}attention layers from 6 to 3 decreases computation from 147 to 125\,GFLOPs and parameters from 44.0M to 41.8M, while PQ drops only from 39.6$\%$ to 38.7$\%$. Further reduction to a single layer significantly impacts PQ (36.9$\%$), indicating that a small number of fusion layers is sufficient but cannot be removed entirely.

\begin{table}[h]
\centering
\caption{Ablation study on the number of Transformer encoder layers in Mask2Former-R50.}
\resizebox{0.47\textwidth}{!}{
\begin{tabular}{lccccc}
\toprule
\textbf{Enc. Layers} & \textbf{PQ(\%)} & \textbf{mIoU(\%)} & \textbf{AP(\%)} & \textbf{Params (M)} & \textbf{GFLOPs} \\
\midrule
6 & 39.6 & 46.0 & 26.5 & 44.0 & 147\\
5 & 39.6 & 45.9 & 26.6 & 43.26 & 143 \\
4 & 38.4 & 46.3 & 25.3 & 42.52 & 134 \\
3 & 38.7 & 44.5 & 24.7 & 41.79 & 125 \\
2 & 38.3 & 46.1 & 24.5 & 41.06 & 116 \\
1 & 36.9 & 46.3 & 23.2 & 40.33 & 107 \\
\bottomrule
\end{tabular}
}
\label{tab:enc_layers_ablation}
\end{table}

Table~\ref{tab:dims_ablation} evaluates channel width reduction. Shrinking the pixel decoder and mask branch from 256 to 128 channels reduces computation from 147 to 97.7\,GFLOPs and parameters from 44.0M to 40.7M, with a limited PQ decrease (39.6$\%$ to 38.0$\%$). Reducing to 64 channels lowers compute further (76.6\,GFLOPs) but causes a larger PQ drop (35.9$\%$), identifying 128 channels as a practical efficiency-accuracy operating point.

\begin{table}[h]
\centering
\caption{Ablation study on the width of the pixel decoder and mask branch in Mask2Former-R50.}
\resizebox{0.47\textwidth}{!}{
\begin{tabular}{lccccc}
\toprule
\textbf{Channel Dims} & \textbf{PQ(\%)} & \textbf{mIoU(\%)} & \textbf{AP(\%)} & \textbf{Params (M)} & \textbf{GFLOPs} \\
\midrule
256 & 39.6 & 46.0 & 26.5 & 44.0 & 147\\
128 & 38.0 & 46.0 & 25.6 & 40.7 & 97.7 \\
64 & 35.9 & 44.3 & 23.4 & 39.2 & 76.6 \\
\bottomrule
\end{tabular}
}
\label{tab:dims_ablation}
\end{table}

\subsubsection{\texorpdfstring{Low{-}Power Embedded Evaluation}{Low-Power Embedded Evaluation}}

Table~\ref{tab:orin15w} reports inference in the Jetson AGX Orin 15\,W mode, which is closer to a constrained deployment regime than MaxN. Compared with Mask2Former{-}R50, LiPS consistently lowers latency while preserving most of the panoptic quality. On ADE20K, LiPS (2 levels) reduces latency from 982\,ms to 651\,ms, while LiPS (full) reaches 814\,ms. On Cityscapes, latency decreases from 1135\,ms for Mask2Former{-}R50 to 953\,ms and 883\,ms for LiPS (full) and LiPS (2 levels), respectively. These gains are accompanied by higher embedded throughput, with LiPS (2 levels) achieving 1.54\,FPS on ADE20K and 1.13\,FPS on Cityscapes, compared with 1.02 and 0.88\,FPS for Mask2Former{-}R50.

The memory measurements further support the intended design.
All LiPS variants also significantly reduce GPU memory usage across both datasets. On ADE20K, peak GPU memory decreases from 2466\,MB for Mask2Former{-}R50 to 1419\,MB ($-42.5\%$) for LiPS (2 levels) and 1634\,MB  ($-33.7\%$)  for LiPS (full). On Cityscapes, it decreases from 3430\,MB to 1991\,MB ($-41.9\%$)  and 2264\,MB ($-34.0\%$), respectively. System RAM varies less due to runtime and framework overheads, whereas GPU memory more clearly reflects the reduction in dense feature activations. These results are consistent with the stage{-}wise analysis and confirm the improved deployment efficiency of LiPS under both latency and memory constraints.

\begin{table}[h]
\centering
\caption{Jetson AGX Orin measurements in 15\,W mode. RAM and GPU memory denote peak inference usage.}
\resizebox{0.47\textwidth}{!}{

\begin{tabular}{lcccc}
\toprule
\textbf{Model} & \textbf{Lat. (ms)} & \textbf{FPS} & \textbf{RAM (MB)} & \textbf{GPU Mem. (MB)} \\
\midrule

\multicolumn{5}{c}{\textbf{ADE20K}} \\
\midrule
Mask2Former-R50 & 982  & 1.02 & 19288 & 2466 \\
LiPS (1 level)  & 630  & 1.59 & 19155 & 1386 \\
LiPS (2 levels) & 651  & 1.54 & 19201 & 1419 \\
LiPS (3 levels) & 729  & 1.37 & 19214 & 1470 \\
LiPS (full)     & 814  & 1.23 & 19231 & 1634 \\

\midrule
\multicolumn{5}{c}{\textbf{Cityscapes}} \\
\midrule
Mask2Former-R50         & 1135 & 0.88 & 20556 & 3430 \\
LiPS (1 level)  & 873  & 1.15 & 20214 & 1946 \\
LiPS (2 levels) & 883  & 1.13 & 20216 & 1991 \\
LiPS (3 levels) & 899  & 1.11 & 20285 & 2057 \\
LiPS (full)     & 953  & 1.05 & 20364 & 2264 \\

\bottomrule
\end{tabular}

}
\label{tab:orin15w}
\end{table}

\section{Discussion}

\begin{figure*}[t]
    \centering
    \includegraphics[width=\textwidth]{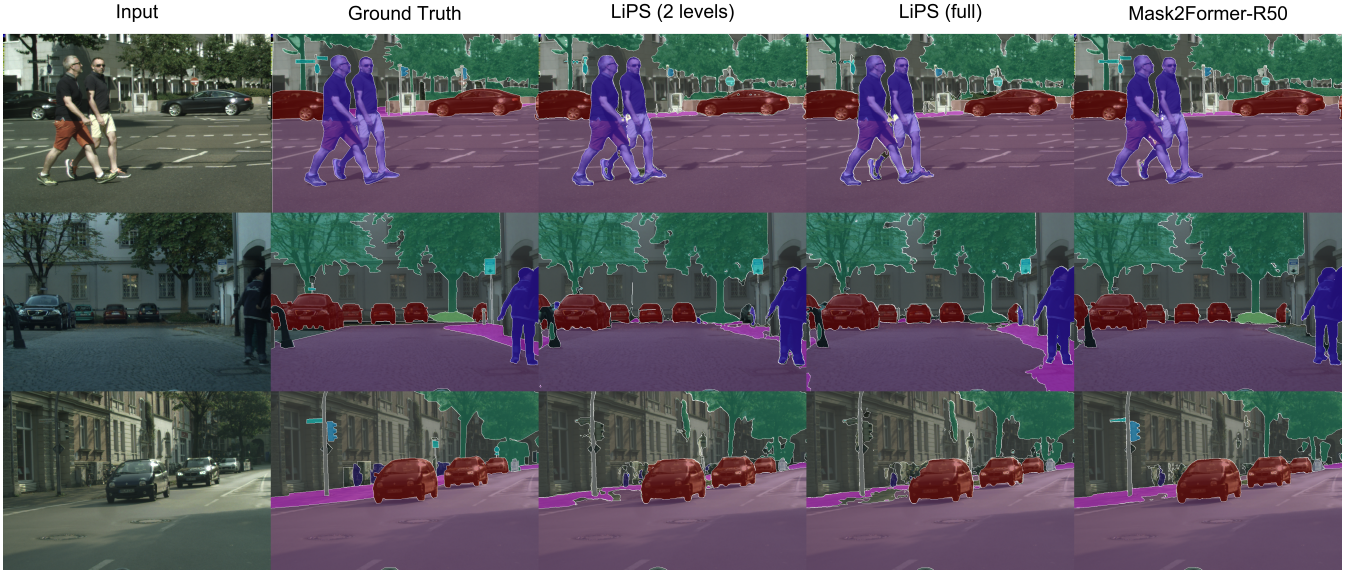}
    \caption{Qualitative comparison on Cityscapes
    Columns (left$\to$right): (a) input, (b) ground truth,
    (c) LiPS (2 levels), (d) LiPS (full),
    (e) Mask2Former{-}R50.
    A fixed ROI is cropped and magnified identically across methods to enable like{-}for{-}like
    inspection of boundary adherence (e.g., curb/sidewalk), thin{-}structure fidelity (poles, traffic signs),
    and stuff/thing separation.
    }
    \label{fig:cityscapes-qual-zoom}
\end{figure*}

LiPS attains most of its efficiency by compressing feature maps before attention. Strided downsampling of routed levels reduces the number of spatial positions processed by multi{-}scale deformable attention, while a shallow fusion stack and lightweight top{-}down path concentrate savings in the dominant upstream stages and leave the query decoder unchanged.

LiPS is complementary to deployment optimizations such as pruning, distillation, and quantization. Since all reported models, including Mask2Former{-}R50, are evaluated using the same TensorRT framework with FP16 (half{-}precision) inference, the observed gains cannot be attributed solely to deployment optimization. A quantized Mask2Former would still process the same upstream feature pyramid, whereas LiPS reduces the number of feature tokens reaching the costly fusion stages. LiPS can therefore be viewed as a structural efficiency layer that complements lower{-}level optimization technique.

Compared with lightweight Transformer segmentation architectures LiPS occupies a distinct efficiency regime. While preserving a query-based panoptic formulation, LiPS requires only 26.4-36.8 GFLOPs on ADE20K at 640$\times$640 resolution and achieves up to 45.3\% mIoU. In comparison, Segmenter~\cite{strudel2021segmenter} reports 39.0, 40.6, 43.1, and 50.7 mIoU on ADE20K for Tiny, Small, Base, and Large variants, while operating on semantic segmentation only. \\

LiPS mainly impacts instance AP on small or thin objects, while PQ and mIoU remain comparatively stable. This suggests that LiPS preserves global panoptic and semantic consistency better than fine-grained instance details. As a result, LiPS is less suitable for applications where precise instance boundaries are critical. Instead, LiPS targets robotic perception under strict power, latency, and memory constraints, such as dynamic SLAM, semantic mapping, traversability analysis, or object-aware navigation, where a temporally frequent panoptic signal can be preferable to a stronger model that runs too slowly or exceeds GPU memory limits.

\section{Conclusion}

We presented LiPS, a lightweight query{-}based panoptic segmentation framework for resource{-}constrained robotic perception. Guided by stage{-}wise profiling, LiPS preserves the masked transformer decoder and simplifies the costly upstream feature pathway through feature routing, spatial compression, and a shallow reduced{-}width pixel decoder. Across ADE20K and Cityscapes, LiPS substantially reduces GFLOPs, latency, and GPU memory while preserving most panoptic and semantic performance. Its main limitation is reduced instance AP on small and thin objects, making it best suited to applications where throughput, memory footprint, and semantic consistency are more critical than fine instance boundaries.

\section*{Acknowledgments}
This work was partially funded by the French Defence Innovation
Agency (Agence de l'innovation de d\'efense, AID).

\vfill\pagebreak

\bibliographystyle{IEEEbib}
\bibliography{strings,refs,cvpr}

\end{document}